\documentclass[11pt]{article}

\usepackage{alltt}
\usepackage{amssymb}
\usepackage{latexsym}

\long\def\IGNORE#1\ENDIGNORE{\begingroup \endgroup}

\def\w#1{{\it #1}} 

\def\i{{\mbox{\(^{-1}\)}}}
\def\al{{\mbox{\(\alpha\)}}}
\def\be{{\mbox{\(\beta\)}}}

\def\la{{\mbox{\(\lambda\!\)}}}

\def\union{\cup}

\def\myem{\sc}
\def\implies{\Rightarrow}

\def\freegroup#1{{F(#1)}}
\def\nm#1{{\mbox{NM}(#1)}}
\def\ng#1{{\mbox{NG}(#1)}}

\def\conceptdef#1{{\sc #1}}
\def\myequiv{\Leftrightarrow}

\def\maybelinebreak{}

\newenvironment{remark}[1]{\smallskip{\itshape Remark{#1}.}}{\smallskip}

\def\fg{{\mbox{$\freegroup{V}$}}}

\def\sub#1{{\mbox{\(_{#1}\)}}} 

\def\rra{{\mbox{\(\:\mapsto\:\)}}} 

\def\ra{{\mbox{\(\:\rightarrow\:\)}}} 

\def\preorder{{\mbox{\(\:\rightarrow\:\)}}}

\def\garrow{{\mbox{\(\,\rightharpoonup\,\)}}} 
\def\parrow{{\mbox{\(\,\rightharpoondown\,\)}}} 

\def\squareforqed{\hbox{\rlap{$\sqcap$}$\sqcup$}}
\def\qed{\ifmmode\squareforqed\else{\unskip\nobreak\hfil
\penalty50\hskip1em\null\nobreak\hfil\squareforqed
\parfillskip=0pt\finalhyphendemerits=0\endgraf}\fi}

\def\x{{\it x}}
\def\y{{\it y}}
\def\z{{\it z}}

\title{Group Theory and Grammatical Description}

\author{{\bf Marc Dymetman}\\
  Xerox Research Centre Europe\\
  6, chemin de Maupertuis\\
  38240 Meylan, France\\
  {\tt Marc.Dymetman@grenoble.xrce.xerox.com}} 

\date{January 1998}

\begin{document}

\maketitle
\begin{abstract}
This paper presents a model for linguistic description based
  on group theory. A grammar in this model, or {\em G-grammar} is a
  collection of lexical expressions which are products of logical
  forms, phonological forms, and their inverses. Phrasal
  descriptions are obtained by forming products of lexical expressions
  and by cancelling contiguous elements which are inverses of each
  other. We show applications of this model to parsing and generation,
  long-distance movement, and quantifier scoping. We believe that by
  moving from the free monoid over a vocabulary V --- standard in
  formal language studies --- to the free group over V, deep
  affinities between linguistic phenomena and classical algebra come
  to the surface, and that the consequences of tapping the
  mathematical connections thus established could be considerable.
\end{abstract}

\bibliographystyle{plain}

\section{Introduction}

There is currently much interest in bringing together the tradition of
categorial grammar, and especially the Lambek calculus \cite{Lambek58}, with the more
recent paradigm of linear logic \cite{Girard:87} to which it has strong ties. One
active research area concerns the design of non-commutative versions
of linear logic \cite{Abrusci:91,Retore:93} which can be sensitive to word
order while retaining the hypothetical reasoning capabilities of
standard (commutative) linear logic that make it so well-adapted to
handling such phenomena as quantifier scoping \cite{Dalrymple:95}.

Some connections between the Lambek calculus and group structure have
long been known \cite{Benthem:86}, and linear logic itself has some
aspects strongly reminiscent of groups (the producer/consumer duality
of a formula $A$ with its linear negation $A^{\bot}$), but no serious
attempt has been made so far to base a theory of linguistic description
solely on group structure.

This paper presents such a model, {\em G-grammars} (for ``group
grammars''), and argues that:
\begin{itemize}
\item The standard group-theoretic notion of {\em conjugacy}, which is
  central in G-grammars, is well-suited to a uniform description of
  commutative and non-commutative aspects of language;
\item The use of conjugacy provides an elegant approach to long-distance
  dependency and scoping phenomena, both in parsing and in generation;
\item G-grammars give a symmetrical account of the semantics-phonology
  relation, from which it is easy to extract, via simple group
  calculations, rewriting systems computing this relation for the
  parsing and generation modes.
\end{itemize}

The paper is organized as follows. In section \ref{Group Computation}
we introduce a ``group computation'' model, using standard algebraic
tools such as {\em free groups, conjugacy} and {\em normal subsets}.
The main deviation from traditional mathematical practice is in the
focus given to the notions of {\em compatible preorder} and {\em
  normal submonoid}, whereas those of {\em compatible
  equivalence relation} and {\em normal subgroup} are more usual in
algebra.  Section \ref{G-Grammars} applies this model to linguistic
description, and presents a G-grammar for a fragment of English
involving quantification and relative pronouns. The next two sections
are concerned with generation and parsing, which correspond to two ways
of exploiting the relation of {\em preorder} associated with the
G-grammar, one (generation) in which logical forms are iteratively
rewritten as combinations of logical forms and phonological forms
until only phonological forms are left, the other (parsing) in which
phonological forms are rewritten as combinations of logical forms and
inverses of those until, after cancellation of adjacent inverses,
exactly one logical form is left. Section \ref{Complements} briefly
mentions certain aspects of group computation which cannot be detailed
in the paper: the use of conjugacy for mixing commutative and
non-commutative phenomena, 
the simulation of logic programs, decidability conditions for parsing
and generation, and differences between G-grammars and categorial
grammars.

\section{Group Computation}\label{Group Computation}

A \conceptdef{monoid} $M$ is a set $M$ together with a product $M
\times M \ra M$, written $(a, b) \rra ab$, such that:

\begin{itemize}
\item This product is associative;
\item There is an element $1 \in M$ (the neutral element) with $1a =
  a1 = a$ for all $a \in M$.
\end{itemize}

A \conceptdef{group} is a monoid in which every element $a$ has an
inverse $a\i$ such that $a\i a = a a\i = 1$.

A \conceptdef{preorder} on a set is a reflexive and transitive 
relation
on this set. When the relation is also symmetrical, that
is, $R(x,y) \implies R(y,x)$, then the preorder is called
an \conceptdef{equivalence relation}. When it is antisymmetrical, that
is that is, $R(x,y) \land R(y,x) \implies x = y$, it is
called a \conceptdef{partial order}.

A preorder $R$ on a group $G$ will be said to be
\conceptdef{compatible} with the group product iff, whenever $R(x,y)$ and
$R(x',y')$, then $R(xx',yy')$.

\paragraph{Normal submonoids of a group.} 
We consider a compatible preorder notated $x \preorder y$ on a group
$G$. The following properties, for any $x,y \in G$, are immediate:
\begin{eqnarray*}
x \preorder y &\myequiv& x y\i \preorder 1;\\
x \preorder y &\myequiv& y\i \preorder x\i;\\
x \preorder 1 &\myequiv& 1 \preorder x\i;\\
x \preorder 1 &\implies& y x y\i \preorder 1,\; \mbox{for any}\ y \in G.
\end{eqnarray*}
Two elements $x,x'$ in a group $G$ are said to be
\conceptdef{conjugate} if there exists $y\in G$ such that $x' =  y x y\i$.
The fourth property above says that the set $M$ of elements $x \in G$
such that $x \preorder 1$ is a set which contains along with an
element all its conjugates, that is, a \conceptdef{normal} subset of $G$. As
$M$ is clearly a submonoid of $G$, it will be called a
\conceptdef{normal submonoid} of $G$.

Conversely, it is easy to show that with any normal
submonoid $M$ of $G$ one can associate a preorder
compatible with $G$. Indeed let's define $x \preorder y$ as $x
y\i \in M$. The relation $\preorder$ is clearly reflexive and
transitive, hence is a preorder. It is also compatible
with $G$, for if $x_1 \preorder y_1$ and $x_2 \preorder y_2$, then
$x_1 {y_1}\i$, $x_2 {y_2}\i$ and 
$y_1 (x_2 {y_2}\i) {y_1}\i$
are in $M$; hence 
$x_1 x_2 {y_2}\i {y_1}\i
=
x_1 {y_1}\i y_1 x_2 {y_2}\i {y_1}\i$ 
is in $M$,
implying that $x_1 x_2 \preorder y_1 y_2$, that is, that the
preorder is compatible.

\begin{remark}{}
  In general $M$ is not a subgroup of $G$.  It is iff $x \preorder y$
  implies $y \preorder x$, that is, if the compatible preorder
  $\preorder$ is an equivalence relation (and, therefore, a
  \conceptdef{congruence}) on $G$. When this is the case, $M$ is a
  \conceptdef{normal subgroup} of $G$. This notion plays a pivotal
  role in classical algebra. Its generalization to
  {\em submonoids} of $G$ is basic for the algebraic theory of
  computation presented here.
\end{remark}

If $S$ is a subset of $G$, the intersection of all normal submonoids
of $G$ containing $S$ (resp.\ of all subgroups of $G$ containing $S$)
is a normal submonoid of $G$ (resp.\ a normal subgroup of $G$) and is
called the \conceptdef{normal submonoid closure} $\nm{S}$ of $S$ in
$G$ (resp.\ the \conceptdef{normal subgroup closure} $\ng{S}$ of $S$
in $G$).

\paragraph{The free group over V.} 
We now consider an arbitrary set $V$, called the
\conceptdef{vocabulary}, and we form the so-called {\myem set of atoms
  on $V$}, which is notated $V \union V\i$ and is obtained by taking
both elements $v$ in $V$ and the formal inverses $v\i$ of these
elements.

We now consider the set $\freegroup{V}$ consisting of the empty
string, notated $1$, and of strings of the form $x_1 x_2 ... x_n$,
where $x_i$ is an atom on $V$. It is assumed that such a string is
\conceptdef{reduced}, that is, never contains two consecutive atoms
which are inverse of each other: no substring $v v\i$ or $v\i v$ is
allowed to appear in a reduced string.

When $\alpha$ and $\beta$ are two reduced strings, their
concatenation $\alpha \beta$ can be reduced by eliminating
all substrings of the form $v v\i$ or $v\i v$. It can be
proven that the reduced string $\gamma$ obtained in this
way is independent of the order of such eliminations. In
this way, a product on $\freegroup{V}$ is defined, and it
is easily shown that $\freegroup{V}$ becomes a 
(non-commutative) group, called the \conceptdef{free group} over
$V$ \cite{Hungerford:74}.

\def\fg{{\mbox{$\freegroup{V}$}}}

\paragraph{Group computation.} 
We will say that an ordered pair
$GCS = (V,R)$ is a \conceptdef{group computation structure} if:
\begin{enumerate}
\item $V$ is a set, called the \conceptdef{vocabulary}, or the set of
  \conceptdef{generators}
\item $R$ is a subset of $\fg$, called the \conceptdef{lexicon}, or
  the set of \conceptdef{relators}.\footnote{For readers familiar with
    group theory, this terminology will evoke the classical notion of group
    \conceptdef{presentation} through generators and relators. The
    main difference with our definition is that, in the classical case,
    the set of relators is taken to be symmetrical, that is, to
    contain $r\i$ if it contains $r$. When this additional assumption is made,
    our preorder becomes an equivalence relation.}
\end{enumerate}

The submonoid closure $\nm{R}$ of $R$ in $\fg$ is called
the \conceptdef{result  monoid} of the group computation structure
$GCS$. The elements of $\nm{R}$ will be called \conceptdef{computation
results}, or simply \conceptdef{results}.

If $r$ is a relator, and if $\alpha$ is an arbitrary
element of $\fg$, then $\alpha r \alpha\i$ will be called
a \conceptdef{quasi-relator} of the group computation structure. It is
easily seen that the set $R_N$ of quasi-relators is equal
to the normal subset closure of $R$ in $\fg$, and that
$\nm{R_N}$ is equal to $\nm{R}$.

A \conceptdef{computation} relative to $GCS$ is a finite sequence $c = 
(r_1,
\ldots, r_n)$ of quasi-relators. The product $r_1 \cdots r_n$ in $\fg$
is evidently a result, and is called the \conceptdef{result of the 
computation}
$c$. It can be shown that the result monoid is
entirely covered in this way: each result is the result of
some computation. A computation can thus be seen as a ``witness'', or 
as
a ``proof'', of the fact that a given element of $\fg$ is a result of
the computation structure.\footnote{The analogy with the view in
  constructive logics is clear. There what we call a result is called
  a {\em formula} or a {\em type}, and what we call a computation is called a
  {\em proof}.}

For specific computation tasks, one focusses on results of a certain
sort, for instance results which express a relationship of input-output,
where input and output are assumed to belong to certain object types.
For example, in computational linguistics, one is often interested in
results which express a relationship between a fixed semantic input and a
possible textual output (generation mode) or conversely in results
which express a relationship between a fixed textual input and a
possible semantic output (parsing mode).

If $GCS = (V,R)$ is a group computation structure, and if
$A$ is a given subset of $\fg$, then we will call the pair
$GCSA = (GCS, A)$ a \conceptdef{group computation structure with
acceptors}. We will say that $A$ is the set of acceptors,
or the \conceptdef{public interface}, of $GCSA$. A result of $GCS$
which belongs to the public interface will be called a
\conceptdef{public result} of $GCSA$.

\section{G-Grammars}\label{G-Grammars}

We will now show how the formal concepts introduced above
can be applied to the problems of grammatical description
and computation. We start by introducing a grammar, which
we will call a \conceptdef{G-Grammar} (for ``Group Grammar''), for a
fragment of English (see Fig. 1).

\begin{figure}[t]
  \begin{center}
\begin{alltt}
j \w{john}\i
l \w{louise}\i
p \w{paris}\i
m \w{man}\i
w \w{woman}\i
A\i r(A) \w{ran}\i
A\i s(A,B) B\i \w{saw}\i
E\i i(E,A) A\i \w{in}\i
t(N) N\i \w{the}\i
\al ev(N,X,P[X]) P[X]\i \al\i X N\i \w{every}\i
\al sm(N,X,P[X]) P[X]\i \al\i X N\i \w{some}\i
N\i tt(N,X,P[X]) P[X]\i \al\i X \al \w{that}\i
\end{alltt}    
    \caption{A G-grammar for a fragment of English}
    \label{fig:g-grammar}
  \end{center}
\end{figure}

A G-grammar is a group computation structure with
acceptors over a vocabulary $V = V_{log} \union V_{phon}$
consisting of a set of logical forms $V_{log}$ and a
disjoint set of phonological elements (in the example,
words) $V_{phon}$. Examples of phonological elements are
{\tt \w{john}}, {\tt \w{saw}}, {\tt \w{every}}, examples
of logical forms {\tt j},\, {\tt s(j,l)},\, {\tt
ev(m,\x,sm(w,\y,s(\x,\y)))}; these logical forms
can be glossed respectively as ``john'', ``john saw louise'' and ``for
every man \x, for some woman \y, \x\ saw \y''.

The grammar lexicon, or set of relators, $R$ is given as a list of
``lexical schemes''. An example is given in Fig. 1.  Each line is a
lexical scheme and represents a set of relators in $\fg$. The first
line is a ground scheme, which corresponds to the single relator
{\tt j \w{john}\i}, and so are the next four lines. The fifth
line is a non-ground scheme, which corresponds to an infinite set of
relators, obtained by instanciating the {\em term meta-variable} {\tt A}
(notated in uppercase) to a logical form. So
are the remaining lines. 
We use Greek letters for {\em expression meta-variables} such as {\tt
  \al}, which can be replaced by an arbitrary expression of $\fg$; thus,
whereas the term meta-variables {\tt A}, {\tt B}, ..., range over
logical forms, the expression meta-variables {\tt \al}, {\tt \be},
..., range over products of logical forms and phonological elements
(or their inverses) in $\fg$.\footnote{Expression
  meta-variables are employed in the grammar for forming the set
  of conjugates $\al\,\,{exp}\,\,\al\i$ of certain expressions ${exp}$ (in
  our example, ${exp}$ is {\tt ev(N,X,P[X]) P[X]\i}, {\tt sm(N,X,P[X]) P[X]\i},
  or {\tt X}). Conjugacy allows the enclosed material ${exp}$ to move
  {\em as a
  block} in expressions of \fg, see sections 3.\ and 4.}

The notation {\tt P[\x]} is employed to express the fact that a
logical form containing an {\em argument identifier}\  {\tt \x} is equal to the
application of the abstraction {\tt P} to {\tt \x}. The meta-variable
{\tt X} in {\tt P[X]} ranges over such identifiers ({\tt \x}, {\tt
  \y}, {\tt \z}, ...), which are notated in lower-case italics (and are always
ground). The meta-variable {\tt P} ranges over logical
form abstractions missing one argument (for instance {\tt\la\,\z.s(j,\z)}). 
When matching meta-variables in logical forms, we will
allow limited use of higher-order unification. For instance, one can
match {\tt P[X]} to {\tt s(j,\x)} by taking ${\tt P} = {\tt
  \la\,\z.s(j,\z)}$ and ${\tt X} = {\tt \x}$.

The vocabulary and the set of relators that we have just specified
define a group computation structure $GCS = (V,R)$. We will now
describe a set of acceptors $A$ for this computation structure. We
take $A$ to be the set of elements of $\fg$ which are products of the
following form:
$$S \: {W_n}\i {W_{n-1}}\i \ldots {W_1}\i$$
where $S$ is a logical
form ($S$ stands for ``semantics''), and where each $W_i$ is a
phonological element ($W$ stands for ``word''). The expression above
is a way of encoding the ordered pair consisting of the logical form
$S$ and the phonological string $W_1 W_2 \ldots W_n$ (that is, the
inverse of the product ${W_n}\i {W_{n-1}}\i \ldots {W_1}\i$).

A public result $S {W_n}\i {W_{n-1}}\i \ldots {W_1}\i$ in
the group computation structure with acceptors $((V,R),A)$
--- the G-grammar ---will be interpreted as meaning that
the logical form $S$ can be expressed as the phonological
string $W_1 W_2 \ldots W_n$.

Let us give an example of a public result relative to the
grammar of Fig. 1.

\noindent We consider the relators (instanciations of relator schemes):
\begin{alltt}
r\sub1 = j\i s(j,l) l\i \w{saw}\i
r\sub2 = l \w{louise}\i
r\sub3 = j \w{john}\i
\end{alltt}
and the quasi-relators:
\begin{alltt}
r\sub1' = j r\sub1 j\i
r\sub2' = (j \w{saw}) r\sub2 (j \w{saw})\i
r\sub3' = r\sub3
\end{alltt}
Then we have:
\begin{alltt}
r\sub1' r\sub2' r\sub3' =
  j j\i s(j,l) l\i \w{saw}\i j\i \(\cdot\)
    j \w{saw} l \w{louise}\i \w{saw}\i j\i \(\cdot\)
      j \w{john}\i = s(j,l) \w{louise}\i \w{saw}\i \w{john}\i
\end{alltt}
which means that {\tt s(j,l) \w{louise}\i \w{saw}\i \w{john}\i} is the
result of a computation {\tt (r\sub1',r\sub2',r\sub3')}. This result
is obviously a public one, which means that the logical form {\tt
  s(j,l)} can be verbalized as the phonological string \w{john saw
  louise}.

\section{Generation}\label{Generation}

Applying directly, as we have just done, the definition of a group
computation structure in order to obtain public results can be
somewhat unintuitive. It is often easier to use the
preorder $\preorder$. If, for $a,b,c \in \fg$, $abc$ is a relator,
then $abc \preorder 1$, and therefore $b \preorder a\i c\i$. Taking
this remark into account, it is possible to write the relators of our
G-grammar as the ``rewriting rules'' of Fig.\ \ref{fig:gen-rules};
we use the notation $\garrow$ instead of $\preorder$ to distinguish
these rules from the parsing rules which will be introduced in the next section.

\begin{figure}[t]
\begin{center}
\begin{alltt}
j \garrow \w{john}
l \garrow \w{louise}
p \garrow \w{paris}
m \garrow \w{man}
w \garrow \w{woman}
r(A) \garrow A \w{ran}
s(A,B) \garrow A \w{saw} B
i(E,A) \garrow E \w{in} A
t(N) \garrow \w{the} N
ev(N,X,P[X]) \garrow \al\i \w{every} N X\i \al P[X]
sm(N,X,P[X]) \garrow \al\i \w{some} N X\i \al P[X]
tt(N,X,P[X]) \garrow N \w{that} \al\i X\i \al P[X]
\end{alltt}                 
\caption{Generation-oriented rules}
\label{fig:gen-rules}
\end{center}
\end{figure}

The rules of Fig.\ \ref{fig:gen-rules} have a systematic
structure. The left-hand side of each rule consists of a single
logical form, taken from the corresponding relator in the G-grammar;
the right-hand side is obtained by ``moving'' all the remaining
elements in the relator to the right of the arrow.

Because the rules of Fig.\ \ref{fig:gen-rules} privilege the rewriting
of a logical form into an expression of \fg, they are called {\em
  generation-oriented rules} associated with the G-grammar.

Using these rules, and the fact that the preorder $\garrow$ is
compatible with the product of \fg, the fact that {\tt s(j,l)
  \w{louise}\i \w{saw}\i \w{john}\i} is a public result can be
obtained in a simpler way than previously. We have:
\begin{alltt}
s(j,l) \garrow j \w{saw} l
j \garrow \w{john}
l \garrow \w{louise}
\end{alltt}
by the seventh, first and second rules (properly instanciated), and
therefore, by transitivity and compatibility of the preorder:
\begin{alltt}
s(j,l) \garrow j \w{saw} l 
  \garrow \w{john} \w{saw} l \garrow \w{john} \w{saw} \w{louise}
\end{alltt}
which proves that {\tt s(j,l) \garrow  \w{john} \w{saw} \w{louise}},
which is equivalent to saying that {\tt s(j,l)
  \w{louise}\i \w{saw}\i \w{john}\i} is a public result.


\begin{figure}[p]
\begin{center}
\begin{alltt}
i(s(j,l),p)
\garrow s(j,l) \w{in} p
\garrow j \w{saw} l \w{in} p
\garrow \w{john} \w{saw} l \w{in} p
\garrow \w{john} \w{saw} \w{louise} \w{in} p
\garrow \w{john} \w{saw} \w{louise} \w{in} \w{paris}

ev(m,\x,sm(w,\y,s(\x,\y)))
\garrow \al\i \w{every} m \x\i \al sm(w,\y,s(\x,\y))
\garrow \al\i \w{every} m \x\i \al \be\i \w{some} w \y\i \be s(\x,\y)
\garrow \al\i \w{every} \w{man} \x\i \al 
\qquad\qquad\be\i \w{some} \w{woman} \y\i \be \x \w{saw} \y
\garrow \al\i \w{every} \w{man} \x\i \al \x \w{saw} \w{some} \w{woman}
\qquad\rm{(by taking \be\ = \w{saw}\i\,\x\i)}
\garrow \w{every} \w{man} \w{saw} \w{some} \w{woman}
{\qquad{\rm{(by taking \al\ = 1)}}}

\tt{}sm(w,\y,ev(m,\x,s(\x,\y)))
\garrow \be\i \w{some} w \y\i \be ev(m,\x,s(\x,\y)))
\garrow \be\i \w{some} w \y\i \be \al\i \w{every} m \x\i \al s(\x,\y)
\garrow \be\i \w{some} \w{woman} \y\i \be 
\qquad\qquad\al\i \w{every} \w{man} \x\i \al \x \w{saw} \y
\garrow \be\i \w{some} \w{woman} \y\i \be \w{every} \w{man} \w{saw} \y
\qquad{\rm{(by taking \al\ = 1)}}
\garrow \w{every} \w{man} \w{saw} \w{some} \w{woman}
{\qquad{{\rm(by taking \be\ = \w{saw}\i \w{man}\i \w{every}\i)}}}

\tt{}r(t(tt(m,\x,s(l,\x))))
\garrow t(tt(m,\x,s(l,\x))) \w{ran}
\garrow \w{the} tt(m,\x,s(l,\x)) \w{ran}
\garrow \w{the} m \w{that} \al\i \x\i \al s(l,\x) \w{ran}
\garrow \w{the} \w{man} \w{that} \al\i \x\i \al s(l,\x) \w{ran}
\garrow \w{the} \w{man} \w{that} \al\i \x\i \al l \w{saw} \x \w{ran}
\garrow \w{the} \w{man} \w{that} \al\i \x\i \al \w{louise} \w{saw} \x \w{ran}
\garrow \w{the} \w{man} \w{that} \w{louise} \w{saw} \w{ran}
\qquad{\rm(by taking \al\ = \w{saw}\i \w{louise}\i)}
\end{alltt}
\caption{Generation examples}
\label{fig:gen-examples}
\end{center}
\end{figure} 


Some other generation examples are given in Fig.\
\ref{fig:gen-examples}.

The first example is straightforward and works
similarly to the one we have just seen: from the
logical form {\tt i(s(j,l),p)} one can derive
the phonological string \w{john saw louise in paris}. 

\paragraph{Long-distance movement and quantifiers}
The second and third examples are parallel to each other and show the
derivation of the same string \w{every man saw some woman} from two
different logical forms.  The penultimate and last
steps of each example are
the most interesting. 
In the penultimate step of the second example, $\be$ is 
instanciated to {\tt \w{saw}\i\,\x\i}. This has the effect of
``moving'' {\em as a whole} the expression {\tt \w{some} \w{woman}
  \y\i} to the position just before {\tt \y}, and therefore to allow
for the cancellation of {\tt \y\i} and {\tt \y}. The net effect is thus to
``replace" the identifier {\tt \y} by the string {\tt \w{some}
  \w{woman}}; in the last step $\al$ is instanciated to the neutral
element $1$, which has the effect of replacing \x\ by \w{every} \w{man}.
In the penultimate step of the third example, $\al$ is instanciated to the neutral
element, which has the effect of replacing \x\ by \w{every} \w{man};
then $\be$ is instanciated to \w{saw}\i \w{man}\i \w{every}\i, which
has the effect of replacing \y\ by \w{some} \w{woman}.

\begin{remark}{}
  In all cases in which an expression similar to {\tt \al\ a$_1$
    $\ldots$ a$_m$ \al\i} appears (with the {\tt a$_i$} arbitrary
  vocabulary elements), it is easily seen that, by giving {\tt \al} an
  appropriate value in $\fg$, the {\tt a$_1$ $\ldots$ a$_m$} can move
  arbitrarily to the left or to the right, {\em but only together in
    solidarity}; they can also freely permute cyclically, that
  is, by giving an appropriate value to {\tt \al}, the expression {\tt
    \al\ a$_1$ $\ldots$ a$_m$ \al\i} can take on the value
  \maybelinebreak {\tt a$_k$ a$_{k+1}$ $\ldots$ a$_m$ a$_1$ $\ldots$
    a$_{k-1}$} (other permutations are in general not possible).  The
  values given to the {\tt \al}, {\tt \be}, etc., in the examples
  of this paper can be understood intuitively in terms of these two
  properties.
\end{remark}

We see that, by this mechanism of concerted movement,
quantified noun phrases can move to whatever place is assigned to them
after the expansion of their ``scope" predicate, a place which was
unpredictable at the time of the expansion of the quantified logical
form. The identifiers act as ``target markers'' for the quantified
noun phrase: the only way to ``get rid'' of an identifier $\x$ is by
moving $\x\i$, {\em and therefore with it the corresponding quantified noun
phrase}, to a place where it can cancel with $\x$.

The fourth example exploits a similar mechanism for handling relative
clauses. At the time the relative pronoun is produced, an identifier
inverse $\x\i$ is also produced which has the capability of moving to
whatever position is finally assigned to the relative verb's argument
$\x$.

\section{Parsing}\label{Parsing}

To the compatible preorder $\preorder$ on $\fg$ there corresponds a
``reverse" compatible preorder $\parrow$ defined as $a \parrow b$ iff $b \preorder
a$, or, equivalently, $a\i \preorder b\i$. The normal submonoid $M'$ in
$\fg$ associated with $\parrow$ is the inverse monoid of the normal
submonoid $M$ associated with $\preorder$, that is, $M'$ contains $a$ iff
$M$ contains $a\i$.

It is then clear that one can present the relations:
\begin{alltt}
j \w{john}\i{}\preorder 1
A\i{}r(A) \w{ran}\i{}\preorder 1
\al sm(N,X,P[X]) P[X]\i{}\al\i{}X N\i{}\w{some}\i{}\preorder 1
\rm{etc.}
\end{alltt}    
in the equivalent way:
\begin{alltt}
\w{john} j\i{}\parrow 1
\w{ran} r(A)\i{}A \parrow 1
\w{some} N X\i{}\al P[X] sm(N,X,P[X])\i{}\al\i{}\parrow 1
\rm{etc.}
\end{alltt}    

Suppose now that we move to the right of the $\parrow$ arrow all elements 
appearing on the left of 
it, but for the single phonological element
of each relator. 
We obtain the rules of Fig.\ \ref{fig:parse-rules}, which we call the
``parsing-oriented'' rules associated with the G-grammar.

\begin{figure}[t]
\begin{center}
\begin{alltt}
\w{john} \parrow j
\w{louise} \parrow l
\w{paris} \parrow p
\w{man} \parrow m
\w{woman} \parrow w
\w{ran} \parrow A\i r(A)
\w{saw} \parrow A\i s(A,B) B\i
\w{in} \parrow E\i i(E,A) A\i
\w{the} \parrow t(N) N\i
\w{every} \parrow \al ev(N,X,P[X]) P[X]\i \al\i X N\i
\w{some} \parrow \al sm(N,X,P[X]) P[X]\i \al\i X N\i
\w{that} \parrow N\i tt(N,X,P[X]) P[X]\i \al\i X \al
\end{alltt}
\caption{Parsing-oriented rules}
\label{fig:parse-rules}
\end{center}
\end{figure}

By the same reasoning as in the generation case, it is easy to show 
that any derivation using 
these rules and leading to the relation $PS \parrow LF$, where $PS$ is a 
phonological string and 
$LF$ a logical form, corresponds to a public result $LF \: PS\i$ in 
the G-grammar. 

A few parsing examples are given in Fig.\ \ref{fig:parse-examples};
they are the converses of the generation examples given earlier.

\begin{figure*}[p]
\begin{center}
\begin{alltt}
\w{john} \w{saw} \w{louise} \w{in} \w{paris}
\parrow j A\i s(A,B) B\i l E\i i(E,C) C\i p
\parrow s(j,B) B\i l E\i i(E,p)
\parrow s(j,l) E\i i(E,p)
\parrow i(s(j,l),p)

\w{every} \w{man} \w{saw} \w{some} \w{woman}
\parrow \al ev(N,\x,P[\x]) P[\x]\i \al\i \x N\i m A\i s(A,B) B\i 
     \be sm(M,\y,Q[\y]) Q[\y]\i \be\i \y M\i w
\parrow \al ev(m,\x,P[\x]) P[\x]\i \al\i \x A\i s(A,B) B\i \be sm(w,\y,Q[\y]) Q[\y]\i \be\i \y
\parrow \x A\i ev(m,\x,P[\x]) P[\x]\i s(A,B) B\i \be sm(w,\y,Q[\y]) Q[\y]\i \be\i \y
\parrow \x A\i ev(m,\x,P[\x]) P[\x]\i s(A,B) Q[\y]\i sm(w,\y,Q[\y]) B\i \y
\parrow ev(m,\x,P[\x]) P[\x]\i s(\x,\y) Q[\y]\i sm(w,\y,Q[\y])
{\rm and then either:}
  \parrow ev(m,\x,P[\x]) P[\x]\i sm(w,\y,s(\x,\y))
  \parrow ev(m,\x,sm(w,\y,s(\x,\y)))
{\rm or:}
  \parrow ev(m,\x,s(\x,\y)) Q[\y]\i sm(w,\y,Q[\y])
  \parrow sm(w,\y,ev(m,\x,s(\x,\y))

\w{the} \w{man} \w{that} \w{louise} \w{saw} \w{ran}
\parrow t(N) N\i m M\i tt(M,\x,P[\x]) P[\x]\i \al\i \x \al l A\i s(A,B) B\i C\i r(C)
\parrow t(m) M\i tt(M,\x,P[\x]) P[\x]\i \al\i \x \al s(l,B) B\i C\i r(C)
\parrow t(m) M\i tt(M,\x,P[\x]) P[\x]\i s(l,B) \x B\i C\i r(C)
\parrow t(m) M\i tt(M,\x,P[\x]) P[\x]\i s(l,\x) C\i r(C)
\parrow t(m) M\i tt(M,\x,s(l,\x)) C\i r(C)
\parrow tt(t(m),\x,s(l,\x)) C\i r(C)
\parrow r(tt(t(m),\x,s(l,\x)))
\end{alltt}
\caption{Parsing examples}
\label{fig:parse-examples}
\end{center}
\end{figure*}


In the first example, we first rewrite each of the phonological 
elements into the expression 
appearing on the right-hand side of the rules 
(and where the meta-variables have been renamed 
in the standard way to avoid name clashes). The rewriting has taken 
place in parallel, which is 
of course permitted (we could have obtained the same result by 
rewriting the words one by one). 
We then perform certain unifications: {\tt A} is unified with {\tt j}, 
{\tt C} with {\tt p}; 
then {\tt B} is unified to {\tt l}.%
\footnote{Another possibility at this point would be to unify {\tt l}
  with {\tt E} rather than with {\tt B}. This would lead to the construction of the
  logical form {\tt i(l,p)}, and, after unification of {\tt E} with that
  logical form, would conduct to the output {\tt s(j,i(l,p))}. If one
  wants to prevent this output, several approaches are possible. The
  first one consists in typing the logical form with syntactic
  categories. The second one is to have some notion of logical-form
  well-formedness (or perhaps interpretability) disallowing
  the logical forms {\tt i(l,p)} [louise in paris] or {\tt i(t(w),p)} [(the woman) in paris], although it might
  allow the form {\tt t(i(w,p))} [the (woman in paris)].}  
Finally {\tt E} is unified with {\tt s(j,l)}, and we obtain the logical form {\tt i(s(j,l),p)}. In this
last step, it might seem feasible to unify {\tt E} to {\tt i(E,p)}
instead, but that is in fact forbidden for it would mean that the
logical form {\tt i(E,p)} is not a finite tree, as we do require.
This condition prevents ``self-cancellation" of a logical form with a
logical form that it strictly contains.

\paragraph{Quantifier scoping}
In the second example, we start by unifying {\tt m} with {\tt N} and
{\tt w} with {\tt M}; then we ``move" {\tt P[\x]\i} next to {\tt
  s(A,B)} by taking {\tt \al\ = \x\,A\i};\footnote{We have assumed
  that the meta-variables corresponding to identifiers in {\tt P}
  and {\tt Q} have been instanciated to arbitrary, but different,
  values \x\ and \y. See \cite{ThisAuthor98} for a discussion of this point.}
then again we 
``move" {\tt Q[\y]\i} next to {\tt s(A,B)} by taking 
{\tt \be\ = B sm(w,\y,Q[\y])\i}; {\tt 
\x} is then unified with {\tt A} and {\tt \y} with {\tt B}. This leads 
to the expression:
\begin{alltt}
   ev(m,\x,P[\x])P[\x]\i{}s(\x,\y)Q[\y]\i{}sm(w,\y,Q[\y])
\end{alltt}
where we now have a choice. We can either unify {\tt s(\x,\y)} with 
{\tt Q[\y]}, or with {\tt 
P[\x]}. In the first case, we continue by now unifying {\tt P[\x]} with 
{\tt 
sm(w,\y,s(\x,\y))}, leading to the output
{\tt ev(m,\x,sm(w,\y,s(\x,\y)))}. In 
the second case, we continue by now unifying {\tt Q[\y]} with {\tt 
ev(m,\x,s(\x,\y))}, 
leading to the output {\tt sm(w,\y,ev(m,\x,s({\it
    x},\y))}. The 
two possible quantifier 
scopings for the input string are thus obtained, each corresponding to 
a certain order of 
performing the unifications.

In the last example, the most interesting step is the one (third step) in which 
{\tt \al} is instanciated to {\tt s(l,B)\i}, which has the effect of ``moving'' {\tt x} 
close to the ``missing'' 
argument {\tt B\i} of ``louise saw", to cancel it by unification with 
{\tt B} and consequently to 
fill the second argument position in the logical form headed by {\tt 
s}. After this step, 
{\tt P[x]} is ready to be unified with {\tt s(l,x)}, finally 
leading to the expected 
logical form output for the sentence.

\section{Final remarks}\label{Complements}

{\bf Mixing commutative and non-commutative phenomena\quad}
We have already seen examples where commutative and
non-commutative aspects are both present in a lexical entry. Thus, the
presence of $\al,\al\i$ in the entry for `every' in
Fig.~\ref{fig:gen-rules} allows the expression {\tt \w{every} N X\i}
to move as a block to the position where the argument {\tt X} of the
verb is eventually found; in this movement, it is however impossible
for {\tt \w{every}} and {\tt N} to exchange their relative positions,
and it can be shown that, for the input logical forms of
Fig.~\ref{fig:gen-examples}, only the four phonological strings listed can
be obtained.

Let us briefly indicate how this commutative/non-commutative
partnership could be used for error correction purposes. Suppose that a
relator of the form $\al\, {\sf Error}\i\, {\sf Repair}\,\al\i
{\sf Report}$ is added to the grammar, where ${\sf Error}$ is some
erroneous input (for instance it could be the word `principle'
improperly used in a situation where `principal' is needed),
${\sf Repair}$ is what the input should have been (in our example, the
word `principal'), and ${\sf Report}$ is a report which tells us how the
error was corrected. Then, the expression ${\sf Error}\i\,
{\sf Repair}$ can move in block to the spot where the error occurs,
replacing the erroneous word by the correction and allowing normal
processing to continue. The report can
then be used for warning or evaluation purposes.

{\bf Fully commutative structures and logic programs\quad} Suppose that one wants to have a group computation
structure which is completely commutative, that is, one in which
elements can move freely relative to each other. This
property could be stipulated by introducing  a notion of ``commutative group
computation'' using the free commutative group ${FC}(V)$ rather than
$\fg$. Another possibility, which illustrates the flexibility of
the group computation approach, is just to add a
relator scheme $\al\be\al\i\be\i$ to $R$, where $\al$ and $\be$ are
expression meta-variables. This expression is called a {\em commutator} of $\al$
and $\be$ because when multiplied by $\be\al$ it yields $\al\be$. This
single relator scheme permits to permute elements in any expression,
and has the same effect as using ${FC}(V)$. 

An example where commutative structures are useful is the case
of logic programs. It can be shown that if one encodes a clause of a logic program $P_0 \leftarrow
P_1 \ldots P_n$ as a relator scheme $P_0 P_n\i \ldots P_1\i$ in a
commutative structure, and defines
public results to consist of a single ground predicate $P$, then public
results in the group computation structure coincide with consequences
of the program.

{\bf G-grammars, rewriting, and computability\quad} In the discussion
of parsing and generation, we saw how a derivation according to the
rewriting rules of Figs.~\ref{fig:gen-rules} and \ref{fig:parse-rules}
is always ``sound'' with respect to the group computation structure.
We did not consider the opposite question, namely whether it is
``complete'' with respect to it: can any public result relative to the
GCS be obtained by such rewritings? The answer to this question is
given by a theorem demonstrated in \cite{ThisAuthor98}, which states that
such a rewriting system is complete relative to the GCS if the system does not
contain ``ground cycles'', that is situations where a ground term $T$ can
derive $\ldots T \ldots$. This condition is true of both the rewriting
systems of Figs~\ref{fig:gen-rules} and \ref{fig:parse-rules}. For
instance, in the generation case, it can be checked that any ground
logical form that appears on the right-hand side of a rule of
Fig.~\ref{fig:gen-rules} is {\em strictly smaller} than the ground
logical form on the left-hand side, therefore precluding ground
cycles. This condition is related to the {\em computability} of
generation and parsing. Thus it is obvious that the decreasing size
property of generation rules implies that any derivation from a ground
logical form is bounded in length. This can be used to prove
that generation is decidable and can only produce a finite number of
results for a fixed input, and a similar argument can be made for
parsing. In the terminology of \cite{Dy91Inherent} a grammar with this
properties is said to be {\em inherently reversible}; this property is
difficult to guarantee in formalisms relying on empty categories for
long-distance dependencies, a problem which does not appear when using
conjugacy for the same purpose.

{\bf G-grammars, DCGs and categorial grammars\quad} In a similar way
to what was done in the case of a logic program, it is possible to
simulate a definite clause grammar with a group computation
structure. The translation of a clause $A_0 \rightarrow
A_1 \ldots A_n$ becomes the relator scheme $A_0 A_n\i \ldots
A_1\i$. The only difference is that (i) the product is
non-commutative, and (2) the $A_i$'s, for $1\leq i \leq n$, can be
nonterminals or words. The group computation structure obtained
computes the same derivations as the DCG under the condition that the
DCG is not ground-cyclic. This is always the case if the nonterminal
on the left-hand side contains more linguistic material than each
nonterminal on the right-hand side, a natural condition to have in
most cases.

There is however one situation in which this condition is not natural,
namely that of context-free grammars, the limiting case of DCGs where
nonterminals do not contain any linguistic material. Thus, in a CFG,
rules such as: ${VP} \rightarrow {often}\, {VP}$ are standard. Here we
have a ground cycle in the rule itself. Its translation as a relator
is: ${VP}\, {VP}\i\, {often}\i,$ which simplifies to ${often}\i$. The
consequence is that, in the GCS, the word `often' can be added freely
in any string generated from $S$! In order to make the CFG simulable
by a CGS, we have first to enrich it into a DCG which is not
ground-cyclic. This can be easily done by including terms that
memorize derivations or the length of terminal strings.

The context-free rule just discussed can be used to illustrate an
important difference between G-grammars and categorial grammars. In a
categorial grammar, an expression such as ${VP}/{VP}$ is not a neutral
element: it can only combine with a VP to give a VP, not with an NP to
give an NP. By contrast, in a group computation structure, an
expression such as ${VP}\:{VP}\i$ is indistinguishable from the
neutral element, a feature that is required if one wishes to gain access to
many standard mathematical tools.


\newpage

\begin{thebibliography}{1}
\bibitem{Abrusci:91}
V.M. Abrusci.
\newblock Phase semantics and sequent calculus for pure non-commutative
  classical linear logic.
\newblock {\em Journal of Symbolic Logic}, 56(4), 1991.

\bibitem{Dalrymple:95}
M.~Dalrymple, J.~Lamping, F.~Pereira, and V.~Saraswat.
\newblock Linear logic for meaning assembly.
\newblock In {\em Proc. CLNLP}, Edinburgh, 1995.

\bibitem{Dy91Inherent}
Marc Dymetman.
\newblock Inherently reversible grammars, logic programming, and computability.
\newblock In {\em Proceedings of the ACL Workshop: Reversible Grammar in
  Natural Language Processing}, pages 20--30, Berkeley, CA, June 1991.

\bibitem{ThisAuthor98}
Marc Dymetman.
\newblock Group computation and rewriting.
\newblock {\em (in preparation)}, 1998.

\bibitem{Girard:87}
J.Y. Girard.
\newblock Linear logic.
\newblock {\em Theoretical Computer Science}, 50(1), 1987.

\bibitem{Hungerford:74}
Thomas~W. Hungerford.
\newblock {\em Algebra}.
\newblock Springer-Verlag, 1974.

\bibitem{Lambek58}
J.~Lambek.
\newblock The mathematics of sentence structure.
\newblock {\em American Mathematical Monthly}, 65:154--168, 1958.

\bibitem{Retore:93}
C.~R\'etor\'e.
\newblock {\em R\'eseaux et s\'equents ordonn\'es}.
\newblock PhD thesis, Univ. Paris 7, 1993.

\bibitem{Benthem:86}
Johan van Benthem.
\newblock {\em Essays in Logical Semantics}.
\newblock D.~Reidel, Dordrecht, Holland, 1986.
\end{thebibliography}
\end{document}